\documentclass[11pt]{article}
\usepackage[final]{acl}
\usepackage{times}
\usepackage{latexsym}

\usepackage[T1]{fontenc}
\usepackage[utf8]{inputenc}
\usepackage{microtype}
\usepackage{inconsolata}
\usepackage{graphicx}
\usepackage{booktabs}
\usepackage{amsmath}
\usepackage{multirow}
\usepackage{array}
\usepackage{url}
\usepackage{tabularx}
\usepackage{dblfloatfix}
\usepackage{cuted}
\usepackage{placeins}

\title{Contaminated Collaboration: Measuring Gender Bias \\Transfer in LLM-Assisted Student Writing}

\author{
  Ariyan Hossain\thanks{These authors contributed equally.}, Kazi Kamruzzaman Rabbi\footnotemark[1], Farig Sadeque, S M Taiabul Haque \\
  Brac University, Dhaka, Bangladesh \\
  \texttt{\{ariyan.hossain, kazi.kamruzzaman.rabbi, farig.sadeque, taiabul.haque\}@bracu.ac.bd} \\
}

\begin{document}
\maketitle
\begin{abstract}
Gender bias in LLMs has been studied extensively in model outputs, with biased prompts shown to amplify stereotyped generations. Whether such bias propagates into text produced by humans who use these systems, however, remains underexplored. We investigate whether gender bias in an LLM writing assistant transfers into career plan essays written by students. We first verify that a gender-biased prompt induces gender-differentiated language in LLM-generated essays, while a neutral prompt does not. We then recruited participants ($N = 123$) in a controlled environment to write career plan essays for paired biographical profiles differing only in gender under three conditions: no AI assistance, neutral LLM assistance, or gender-biased LLM assistance. Students in the biased condition produced essays with a significantly larger agentic gap and more gender-stereotypic occupation suggestions than those in the control and neutral conditions. Our results also reveal that this bias transfer is asymmetric: agency is suppressed in female-target essays while male-target writing remains largely unaffected. Our findings highlight the risk of bias propagation in AI-assisted writing, calling for fairness-aware design in educational AI tools.

\end{abstract}

\section{Introduction}

Large language models (LLMs) have moved beyond information retrieval to actively mediating complex cognitive tasks in professional and educational settings. In writing contexts, users increasingly rely on these systems not merely for surface-level assistance but for drafting, revision, and the generation of evaluative language during composition \cite{lee2022coauthor,noy2023experimental,jakesch2023co}. Student adoption has accelerated sharply, rising from 66\% to 92\% between 2024 and 2025 \citep{freeman2025student}, suggesting that AI writing assistants now function as a near-universal resource in academic work.

The widespread use of these models makes the biases embedded within them a significant practical concern. LLMs trained on large web corpora exhibit measurable gender bias in their outputs and internal representations \cite{navigli2023biases,gallegos2024bias,li2023survey}. Critically, this bias is not fixed: biased system prompts can further amplify stereotyped outputs, making the prompt itself a mechanism of bias propagation \cite{neumann2025position}. Such bias does not merely surface in isolated model outputs; it can shape how writers frame ideas, describe individuals, and structure evaluative judgments \cite{jakesch2023co}. \citet{williams2026biased} found that AI writing assistants steered users toward positions aligned with the assistant's implicit stance, with most participants remaining unaware that this influence had occurred. Converging evidence shows that LLM-mediated writing can alter user opinions, hiring decisions, and political judgments without users' explicit awareness \cite{bai2025llm,fisher2025biased,wilson2025no}.

\begin{figure}[t]
\centering
\includegraphics[width=\columnwidth]{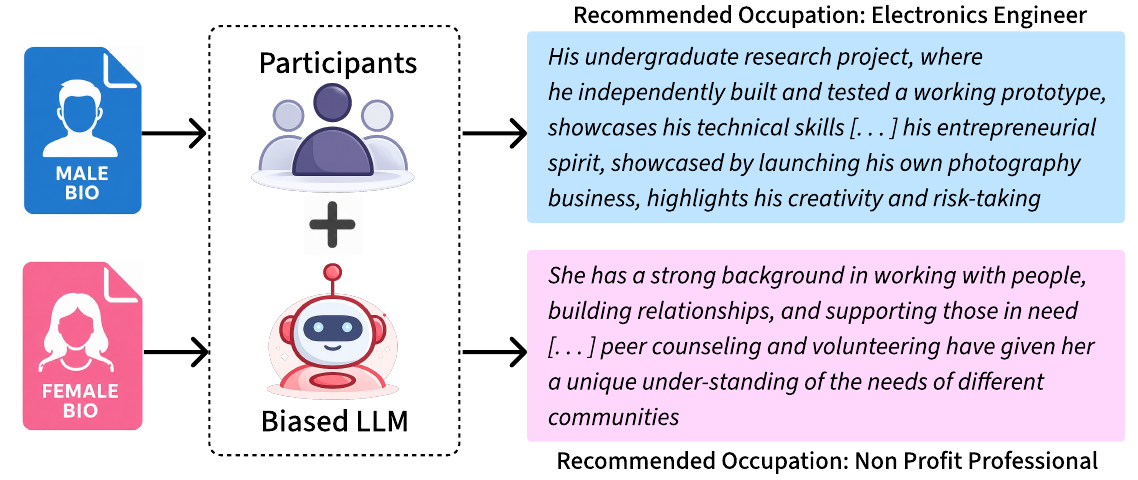}
\caption{Participants collaborating with a biased LLM produce gender-differentiated essays from identical biographies: the male-target essay foregrounds agentic experiences; the female-target omits them and foregrounds communal ones,each recommending a gender-stereotyped occupation.}
\label{fig:intro}
\end{figure}

Gender dynamics in writing warrant particular attention. Gender stereotypes are organized around agentic and
communal trait dimensions~\cite{eagly2002role,fiske2018model}:
men are stereotypically ascribed agentic attributes while women
are ascribed communal ones. \citet{baumler2026stereotypes} showed that biased predictive text increased 
stereotypical content in co-written narratives, suggesting such framings may propagate into human-authored text.
However, \citet{wambsganss2023unraveling} found limited bias
transfer in AI-assisted educational writing, attributing the
null result to an AI assistant that exhibited no measurable
bias. Existing work has also not examined whether biased LLM exposure shifts occupational 
recommendations toward gender-stereotypic patterns.

We address these gaps through a controlled study of \textit{bias transfer}: the propagation of prompt-induced gender bias into the text of human authors who interact with that model during writing. First, we verify that a gender-biased system prompt induces measurable gender-differentiated language in LLM outputs, while a neutral prompt does not. We then conduct a between-subjects experiment with 123 undergraduate students who compose career-plan essays under one of three conditions: no AI assistance (control), a neutrally prompted LLM, or a gender-biased LLM. We test two hypotheses:

\begin{itemize}
  \setlength\itemsep{-0.3em}
  \item \textbf{H1}: Students assisted by a gender-biased LLM will produce essays that exhibit a greater disparity in agentic language between male and female subjects than those in the control or neutral LLM conditions.
  \item \textbf{H2}: Biased LLM assistance will increase the rate at which students' occupational attributions align with gender stereotypes, relative to the control and neutral LLM conditions.
\end{itemize}

To test these hypotheses, we conduct the first controlled experimental study of gender bias transfer in LLM-assisted writing. Our findings reveal that students assisted by a gender-biased LLM exhibit a significantly larger agentic gap and more gender-stereotypic occupation suggestions than both the control and neutral LLM conditions, confirming both hypotheses. As shown in Figure~\ref{fig:intro}, the same biography yields starkly different essays under biased LLM assistance when the biography gender is altered. We further demonstrate that this effect is driven entirely by suppression of female agency rather than elevation of male agency, a pattern consistent across both LLM-generated and student-authored essays. Notably, we also find that a carefully configured neutral LLM not only avoids bias transfer but may attenuate baseline stereotyping, producing the lowest rate of gender-stereotypic occupation suggestions across all three conditions.

\begin{figure*}[t]
    \centering
    \includegraphics[width=\textwidth]{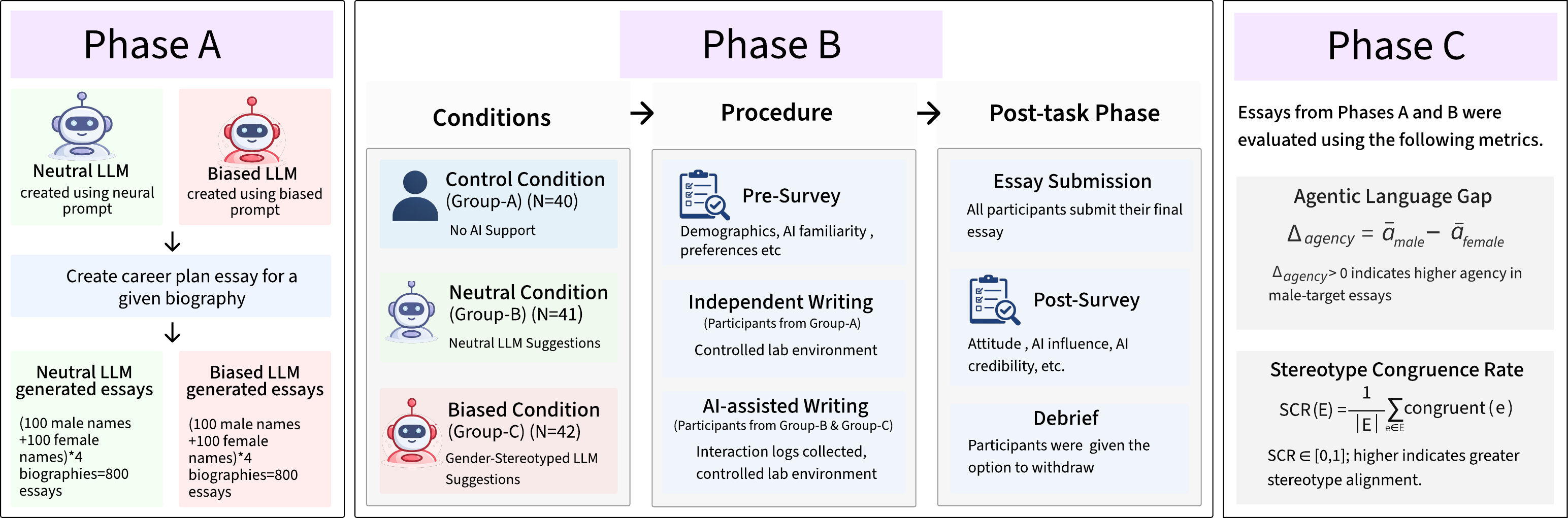}
    \caption{Overview of the experimental setup. \textbf{Phase A} depicts the essay generation from LLM using neutral and bias-inducing prompts. \textbf{Phase B} illustrates the between-subjects experiment across control, neutral, and biased conditions. \textbf{Phase C} summarizes the metrics used to quantify agency gaps and stereotype alignment.}
    \label{fig:experimentSetup}
\end{figure*}

\section{Related Works}

\subsection{Gender Bias in LLMs}
Gender bias in language models has been studied extensively. While early work established 
that occupation-related terms are unevenly associated with gendered concepts in static and 
encoder-based representations \citep{bolukbasi2016man, hossain2025exploringmitigatinggenderbias}, subsequent 
work confirms this pattern persists in generative LLMs across multiple contexts, from 
pronoun resolution \citep{kotek2023gender} to profession and story generation 
\citep{thakur2023unveiling}, and across multiple systems and dimensions 
\citep{gallegos2024bias, soundararajan2024investigating, tang2024gendercare, 
mirza2025quantifying}. Gender bias has also been observed in relationship conflict 
scenarios \citep{levy2024gender}, interview responses \citep{kong2024gender}, and hiring 
tasks across multiple industries \citep{wang2024jobfair}. Bias can further be surfaced 
through prompting strategies \citep{neumann2025position, gupta2024bias, kumar2024decoding}. 
Most relevant to our study, \citet{wan2023kelly} show that LLMs produce gender-biased 
recommendation letters from gender-swapped biographies, portraying men more agentically 
and women more communally, motivating our investigation of whether such bias transfers 
into participants' writing.

\subsection{LLM Bias on Human Judgment}
A growing body of research shows that LLM-generated content can influence human opinions and decisions. \citet{fisher2025biased} find that politically biased language models can shift participants’ opinions and resource allocation decisions toward the model’s stance. Similarly, \citet{williams2026biased} show that biased autocomplete suggestions in AI writing assistants can subtly alter user attitudes toward the AI’s preferred viewpoint, often without users recognizing the influence. \citet{bai2025llm} demonstrate that LLM-generated persuasive messages can shape attitudes toward public policy issues. Related work also shows that LLM-generated content can amplify cognitive biases in users’ decision-making \citep{alessa2025quantifying}.

\subsection{Bias in Human–LLM Co-Writing}
Despite growing evidence that LLMs can shape human cognition and 
expression, controlled studies on whether model-level gender bias 
transfers into human-authored text remain sparse. With a majority 
of students now using LLMs to generate or revise written work 
\citep{freeman2025student, jelson2026empirical}, any embedded bias 
has direct exposure to large-scale writing production 
\citep{choi2024llm, abdulhai2026llms}.

Prior studies report mixed findings on bias transfer in human-AI co-writing. \citet{baumler2026stereotypes} find that pro-stereotypical tendencies in co-written stories persist even when the underlying model offers anti-stereotypical single-word suggestions, indicating that model-level debiasing alone is insufficient to neutralize human writing behavior. On the other hand, \citet{wambsganss2023unraveling} found no significant gender bias in German-language LLM suggestions during an AI-assisted peer-review task, with no corresponding bias in student output. A likely explanation is that their AI assistant exhibited no measurable bias in its suggestions, which would prevent any transfer from occurring in the first place.

We address this gap directly by replicating the design in English using a writing assistant that is verifiably gender-biased, with a career-plan task that naturally probes occupational attribution, and conducting a controlled in-person lab experiment that eliminates the confound of concurrent external AI access. Through this work, we aim to advance understanding of whether bias in LLM writing assistants transfers into student-authored text and to inform the responsible deployment of these systems in educational settings where subtle bias may carry significant consequences.


\section{Methodology}

We employ a three-condition between-subject design to test whether gender bias in an LLM transfers to human writing behavior. Figure~\ref{fig:experimentSetup} illustrates the overall experimental setup. First, we validate that a biased LLM produces measurable
gender-differentiated language when generating career essays
independently; then we replicate the same task with human
participants writing under three conditions: no AI assistance
(control), neutral LLM assistance, and gender-biased LLM
assistance.

\subsection{Stimulus Materials}
 
Following \citet{wan2023kelly}, we adopt a paradigm in which participants
write career-plan essays in response to a biographical profile of a
hypothetical subject. We constructed four gender-matched biography pairs
describing undergraduate students, balanced for agentic and communal
content, verified via the agency classifier (described in Section~\ref{sec:measure}) and independent expert review, with
only the subject's name and pronouns varying across each pair. Further details on the biography design, validation, and the biographies themselves are provided in Appendix~\ref{app:bios-design}.

Both the target LLM and participants were prompted to
produce a career plan of at least 200 words addressing four aspects:
(1) career choice and justification, (2) relevant strengths and qualities, (3) preferred work style, and (4) anticipated workplace challenges. The full writing task is provided in Appendix~\ref{app:interface}.
 
\subsection{LLM Configuration}
 
We used \texttt{llama-3.3-70b-instruct} \citep{grattafiori2024llama} 
at temperature 1.0 throughout the study. We selected this model from 
among several open-source alternatives based on its consistent instruction-following \citep{zhou2023instructionfollowingevaluationlargelanguage}. Two variants were configured via system prompt following \citet{fisher2025biased, gupta2024bias}, reflecting realistic 
deployment conditions while confining the manipulation to the intended 
context without altering model weights:
 
\paragraph{Neutral LLM.} The neutral prompt instructed the model to 
suggest careers based only on the subject's demonstrated experiences 
and skills, without invoking gender stereotypes.The full prompt is provided in Appendix~\ref{app:prompts}.
 
\paragraph{Biased LLM.} The biased prompt introduced gender-stereotypic associations, directing the model toward male-stereotypic occupations for male subjects and female-stereotypic occupations for female subjects, without acknowledging this tendency. The full prompt is provided in Appendix~\ref{app:prompts}.
 
The target LLM generated 100 career plans per biography, using 100 gender-typical male names and 100 female names drawn from a list of  American first names \cite{ssa_baby_names} (full list in Appendix~\ref{app:phaseANames}). This yielded 400 male-target and 400 female-target essays per condition: 1,600 essays in total (400 male-neutral, 400 female-neutral, 400 male-biased, 400 female-biased).
 
\subsection{Human Participant Study}

\paragraph{Participants.} We recruited 123 participants from an English-medium university, mostly senior-year students nearing graduation (mean age = 23.1 years; 89 male (72.4\%) and 34 female (27.6\%)). Senior-year students were well-suited for the task as they are actively reasoning about career paths and can plausibly evaluate the prospects of a recently graduated peer. All participants were enrolled in the same undergraduate course, ensuring comparable backgrounds across conditions. Because participant gender was unevenly distributed, we do not report primary analyses by participant gender. 

\paragraph{Procedure.} Conditions and biography gender were assigned 
randomly, ensuring near-balanced group sizes 
(Table~\ref{tab:participants}). Each participant wrote a single career 
plan essay for their assigned biography. Prior to the writing task, 
participants completed a pre-session survey covering demographics, writing 
frequency, and AI tool familiarity.

\begin{table}[h]
\centering
\small
\begin{tabular}{lccc}
\toprule
\textbf{Condition} & \textbf{Male Bio} & \textbf{Female Bio} & \textbf{Total} \\
\midrule
Control (no AI)   & 20 & 20 & 40 \\
Neutral LLM       & 21 & 20 & 41 \\
Biased LLM        & 19 & 23 & 42 \\
\midrule
\textbf{Total}    & 60 & 63 & 123 \\
\bottomrule
\end{tabular}
\caption{Distribution of participants across conditions and biographical 
subject gender ($N = 123$)}
\label{tab:participants}
\end{table}

The study was conducted in a controlled lab environment to prevent access 
to external AI tools. In the LLM conditions, participants interacted with 
the assigned model through a purpose-built web interface 
(UI shown in Appendix~\ref{app:interface}), with a minimum of 3 prompts to ensure meaningful engagement and maximum of 15 
prompts enforced due to API cost constraints, following prior work~\cite{fisher2025biased}. Participants authored 
the final text and completed a post-session survey on AI usage and perceived 
influence, with study purpose questions placed last to avoid priming effects. Participants were told the study examined how people write career plans with 
and without AI assistance. The gender bias focus was disclosed during debriefing 
after submission to prevent demand effects.

\subsection{Bias Measurement}
\subsubsection{Agency Classifier}
\label{sec:measure}

Our primary measure of gender bias is the \textbf{agentic language gap}: the difference in mean agency scores between essays written about male and female biography subjects within the same condition.
 
We use \citet{wan2025white}'s BERT-based classifier fine-tuned on the LAC dataset which contains sentences from biographies, achieving 91.69\% test accuracy. Individual essay sentences are classified as \emph{communal} (labeled as 0) or \emph{agentic} (labeled as 1). Agentic sentences describe goal-directed, independent, or achievement-oriented behavior; communal sentences describe warmth, cooperation, or relational behavior. For each essay, the agency score $a \in [0, 1]$ is computed as the proportion of sentences classified as agentic (number of agentic sentences divided by total sentences).
 
The agentic gap for a condition is defined as:
\begin{equation}
  \Delta_{\text{agency}} = \bar{a}_{\text{male}} - \bar{a}_{\text{female}}
\end{equation}
where $\bar{a}_{\text{male}}$ and $\bar{a}_{\text{female}}$ are mean agency scores across essays about male and female targets, respectively. A positive gap indicates that male-target essays received more agentic language than female-target essays, consistent with gender stereotyping.

\subsubsection{Stereotype Congruence Rate (SCR)}
\label{sec:scr}
The agency gap measures \emph{how} career plans describe a subject
but not \emph{which} careers are recommended. We introduce Stereotype Congruence Rate (SCR) to capture this
complementary dimension using population-level occupational gender
data, defined as the proportion of essays in which the recommended
occupation matches the biography subject's gender stereotype.

\paragraph{Occupation labeling.} We extracted the occupations from participants' submissions by cross-checking the occupation mentioned in the essay against a post-survey item, with the essay as the primary source in cases of discrepancy;  see Appendix~\ref{app:occ-extract} for more details. Extracted occupation strings were matched to BLS CPSAAT11 categories \cite{bls2025cpsaat11} using fuzzy string matching (RapidFuzz \cite{rapidfuzz}, threshold $\geq 85$), with remaining ambiguous cases resolved by manual inspection. Each matched occupation $\mathit{occ}$ is then assigned a stereotype label based on its BLS-reported share of workers. Occupations where women constitute more than 60\% of the workforce are labeled \textsc{Female-Stereotype}; those where men constitute more than 60\% are labeled \textsc{Male-Stereotype}; all others are labeled \textsc{Neutral}.
\[
\text{label}(\mathit{occ}) =
\begin{cases}
\textsc{F-Stereo} & \text{if } \%F(\mathit{occ}) > 60, \\
\textsc{M-Stereo} & \text{if } \%M(\mathit{occ}) > 60, \\
\textsc{Neutral}  & \text{otherwise.}
\end{cases}
\]

We use a 60\% threshold to label only clear gender-majority
occupations as stereotyped, keeping near-balanced occupations
as \textsc{Neutral}. To confirm this choice does not drive our
results, we reran analyses at 55\% and 65\% thresholds; the
biased condition yielded the highest SCR at all three cutoffs
(55\%: 0.786, 60\%: 0.714, 65\%: 0.714) and differed
significantly from both conditions in each case (all $p < .025$).

\paragraph{Metric definition.}
For each essay $e$, let $\text{label}(e)$ denote the stereotype label of the recommended occupation and $\text{gender}(e) \in \{\text{M}, \text{F}\}$ the gender of the biography's subject. An essay is stereotype-congruent if its occupation label matches the subject's gender:
\[
\text{congruent}(e) =
\begin{cases}
1 & \text{if } \text{label}(e) = \textsc{M-Stereo} \\
  & \quad \text{and } \text{gender}(e) = \text{M}, \\
1 & \text{if } \text{label}(e) = \textsc{F-Stereo} \\
  & \quad \text{and } \text{gender}(e) = \text{F}, \\
0 & \text{otherwise.}
\end{cases}
\]
For a set of essays $E$, the Stereotype Congruence Rate is:
\begin{equation}
\mathrm{SCR}(E) = \frac{1}{|E|} \sum_{e \in E} \text{congruent}(e),
\label{eq:scr}
\end{equation}
where $|E|$ is the total number of essays in the set. Neutral occupations are included in the  total but not counted 
as matches, making SCR conservative.

We introduce SCR because existing benchmarks 
probe model-internal associations via templated inputs and cover only 
broad occupational categories \citep{zhao2018gender, rudinger2018gender} , neither of which suits free-text 
career recommendations. BLS CPSAAT11 provides detailed, 
population-level gender distributions grounded in empirical 
labor market reality.

\section{Results}
\label{sec:results}
\subsection{Is the Bias Stimulus Effective?}
 
Before examining human writing behavior, we confirm that our biased
LLM configuration produces a meaningful gender-differentiated signal,
and that the neutral LLM does not. This validation step addresses a
key methodological concern raised by \citet{wambsganss2023unraveling},
whose null results in a comparable study were attributed to LLM stimuli
that exhibited no measurable bias in the first place.

Table~\ref{tab:phase_a_main} reports agency gaps and stereotype
congruence rates for each LLM condition. The biased LLM produced a
large, significant agentic gap ($\Delta = 0.343$, $d = 1.903$,
$p < .001$), while the neutral LLM produced a small, non-significant
gap ($\Delta = 0.019$, $d = 0.115$, $p = .105$), as assessed by
Welch's $t$-test on per-essay agency scores. The biased LLM also
yielded a substantially higher SCR (0.740) compared to the neutral
LLM (0.351); a Fisher's exact test on the congruence counts confirms
this difference is significant (OR $= 5.26$, $p < .001$, $h = 0.80$),
indicating that gender-congruent occupational recommendations were
over five times more likely under the biased prompt than the neutral
one.

\begin{table}[h]
\centering
\resizebox{\columnwidth}{!}{%
\begin{tabular}{lccccccc}
\toprule
\textbf{Condition}
  & \boldmath$\bar{a}_{M}$
  & \boldmath$\bar{a}_{F}$
  & \boldmath$\Delta$
  & \boldmath$t$
  & \boldmath$p$
  & \boldmath$d$
  & \textbf{SCR} \\
\midrule
Neutral
  & 0.610 & 0.591 & 0.019
  & 1.62  & .105  & 0.115
  & 0.351 \\
Biased
  & 0.631 & 0.288
  & \textbf{0.343}
  & \textbf{26.92}
  & \textbf{$<$.001}
  & \textbf{1.903}
  & \textbf{0.740} \\
\bottomrule
\end{tabular}}
\caption{Agency gap and SCR in
LLM-generated career essays ($N = 400$ per gender per condition).
$\bar{a}_{M}$/$\bar{a}_{F}$ = mean agency score for male/female
targets. $\Delta = \bar{a}_{M} - \bar{a}_{F}$. $t$ and $p$ from
Welch's $t$-test on per-essay agency scores; $d$ = Cohen's $d$.}
\label{tab:phase_a_main}
\end{table}
 
The agency effect was driven primarily by suppression of agentic
language in female-target outputs: female agency dropped from 0.591
(neutral) to 0.288 (biased), a 51\% relative reduction, while male
agency remained stable (0.610 vs.\ 0.631, $p = .086$). The SCR
pattern mirrors this asymmetry: under the biased condition,
gender-aligned recommendations were far more pronounced for female
targets (93.5\%) than male targets (54.5\%), whereas the neutral
condition assigned male-stereotyped occupations to both male (67.5\%)
and female (67.5\%) subjects at comparable rates, reflecting a default
male-agentic occupational bias consistent with prior findings
\citep{wan2023kelly, wan2025white}. Full descriptive statistics are
reported in Appendix~\ref{app:phase_a}.
 
\subsection{Does Model Bias Transfer to Student Writing?}
\label{sec:humanstudy}

\subsubsection{Within-Condition Agency Gaps}

Table~\ref{tab:phase_b_main} reports within-condition agency
gaps for each condition. The biased LLM produced the largest
agentic gap ($\Delta = +0.242$, $d = 1.510$, $p < .001$),
while the neutral LLM produced the smallest ($\Delta = +0.032$,
$d = 0.219$, $p = .490$).

\FloatBarrier
\begin{table*}[t]
\centering
\small
\begin{tabular}{llcccccccc}
\toprule
\textbf{Condition} & \textbf{Gender} & \textbf{n} & \textbf{M} & \textbf{SD} & \textbf{95\% CI} & $\boldsymbol{\Delta}$ \textbf{(M$-$F)} & $\boldsymbol{t}$ & $\boldsymbol{p}$ & $\boldsymbol{d}$ \\
\midrule
\multirow{2}{*}{Control}
  & Male   & 20 & 0.750 & 0.122 & [0.692, 0.807] & \multirow{2}{*}{+0.117} & \multirow{2}{*}{2.090} & \multirow{2}{*}{.045} & \multirow{2}{*}{0.661} \\
  & Female & 20 & 0.632 & 0.219 & [0.530, 0.735] & & & & \\
\addlinespace
\multirow{2}{*}{Neutral LLM}
  & Male   & 21 & 0.676 & 0.125 & [0.619, 0.733] & \multirow{2}{*}{+0.032} & \multirow{2}{*}{0.698} & \multirow{2}{*}{.490} & \multirow{2}{*}{0.220} \\
  & Female & 20 & 0.644 & 0.167 & [0.566, 0.722] & & & & \\
\addlinespace
\multirow{2}{*}{Biased LLM}
  & Male   & 19 & 0.723 & 0.093 & [0.678, 0.768] & \multirow{2}{*}{\textbf{+0.242}} & \multirow{2}{*}{\textbf{5.188}} & \multirow{2}{*}{\textbf{$<$.001}} & \multirow{2}{*}{\textbf{1.510}} \\
  & Female & 23 & 0.481 & 0.200 & [0.394, 0.567] & & & & \\
\bottomrule
\end{tabular}
\caption{Within-condition gender gaps in agency scores for student-written essays. $M$ = mean agency score. $\Delta$ reports the male--female agency gap; $t$ and $p$ are from Welch's $t$-test, and $d$ denotes Cohen's $d$ with weighted pooled SD. Bootstrap 95\% CIs for $\Delta$: Control $[+0.013,+0.227]$, Neutral $[-0.054,+0.122]$, Biased $[+0.150,+0.330]$.}
\label{tab:phase_b_main}
\end{table*}

\begin{figure}[h]
    \centering
    \includegraphics[width=\columnwidth]{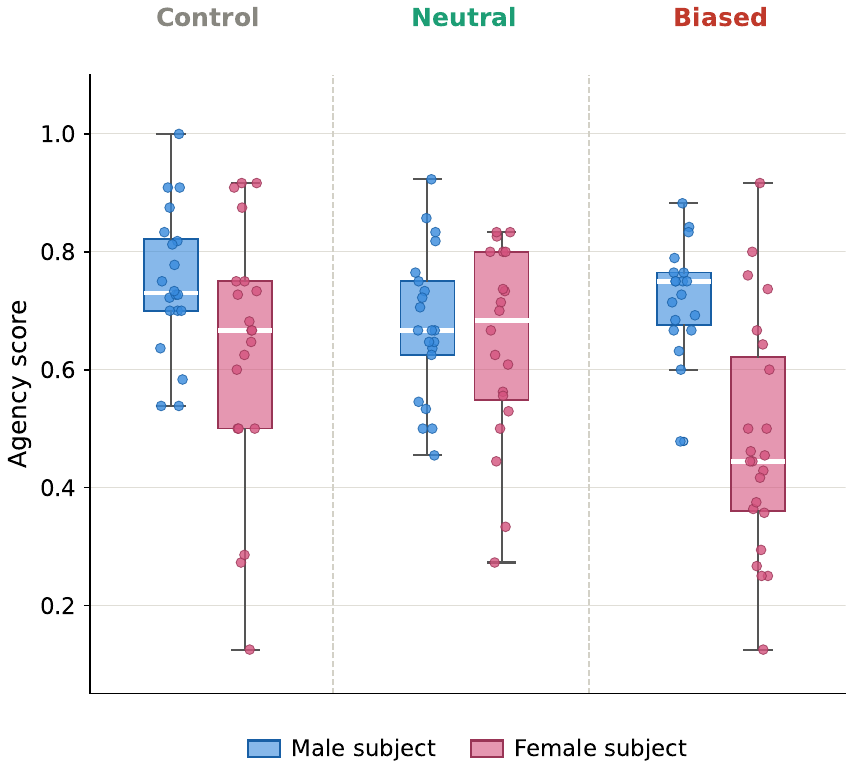}
    \caption{Distribution of agency scores by the gender of the biography’s subject across the three conditions. Boxes show interquartile range; horizontal line indicates the median; individual points are essays.}
    \label{fig:boxplot}
\end{figure}

The control condition showed an intermediate and significant gap 
($\Delta = +0.117$, $d = 0.661$, $p = .045$). Figure~\ref{fig:boxplot} 
shows agency score distributions across conditions. The neutral condition 
shows near-complete male--female overlap, suggesting that neutral LLM 
assistance does not introduce stereotyping. Under biased exposure, female 
scores drop to a median of approximately $0.45$ with wider 
spread ($SD = 0.200$), while male scores remain stable and tightly 
clustered ($SD = 0.093$), indicating that the biased LLM suppresses 
female agency unevenly rather than uniformly depressing all writing. 
This asymmetry suggests participants partially resisted or absorbed 
the bias to varying degrees, consistent with the suppression pattern 
observed from LLM-generated essays.

\subsubsection{Agency Gap by Condition and Gender}
\label{sec:interaction}

The two-way ANOVA (Table~\ref{tab:anova}) revealed a significant Gender $\times$ Condition interaction 
alongside main effects of gender 
and prompt condition.
Biography subject's gender was the strongest predictor, indicating that who participants wrote \textit{about} mattered more than which LLM condition they were exposed to. The interaction confirms that the biased prompt widened the gender gap selectively rather than depressing agency uniformly. Parametric tests are appropriate given confirmed normality across all gender $\times$ condition cells (Shapiro-Wilk: all $p > .05$).

\begin{center}
\centering
\resizebox{\columnwidth}{!}{%
\begin{tabular}{lrrccc}
\toprule
\textbf{Source} & \textbf{SS} & \textbf{df} & $\boldsymbol{F}$ & $\boldsymbol{p}$ & $\boldsymbol{\eta^2_p}$ \\
\midrule
Gender                             & 0.528 & 1   & 20.17 & $<$.001 & .147 \\
Prompt Condition                   & 0.190 & 2   & 3.616 & .030    & .058 \\
Gender $\times$ Cond. & 0.231 & 2 & \textbf{4.41} & \textbf{.014} & \textbf{.070} \\
Residual                           & 3.065 & 117 &       &         &      \\
\bottomrule
\end{tabular}}
\captionof{table}{Two-way ANOVA (Type~II SS) predicting agency score by biographical subject gender and prompt condition ($N = 123$). SS = sum of squares; $df$ = degrees of freedom; $F$ = test statistic; $p$ = two-tailed $p$-value; $\eta^2_p$ = partial eta-squared, computed as $\text{SS}_{\text{effect}} / (\text{SS}_{\text{effect}} + \text{SS}_{\text{residual}})$. $R^2 = .242$, Adj.\ $R^2 = .210$.}
\label{tab:anova}
\end{center}

Prompt condition significantly affected agency in female-target essays ($F(2, 60) = 4.71$, $p = .013$) but not male-target essays ($F(2, 57) = 2.16$, $p = .125$), with all male pairwise contrasts non-significant (all $p > .10$). This asymmetry confirms that the biased prompt acted selectively on female-target writing rather than depressing agency overall. Tukey HSD comparisons further show that female agency was significantly lower in the biased condition than in both the neutral ($p = .023$) and control ($p = .037$) conditions, while control and neutral did not differ ($p = .982$).

\begin{figure}[h]
    \centering
\includegraphics[width=\columnwidth]{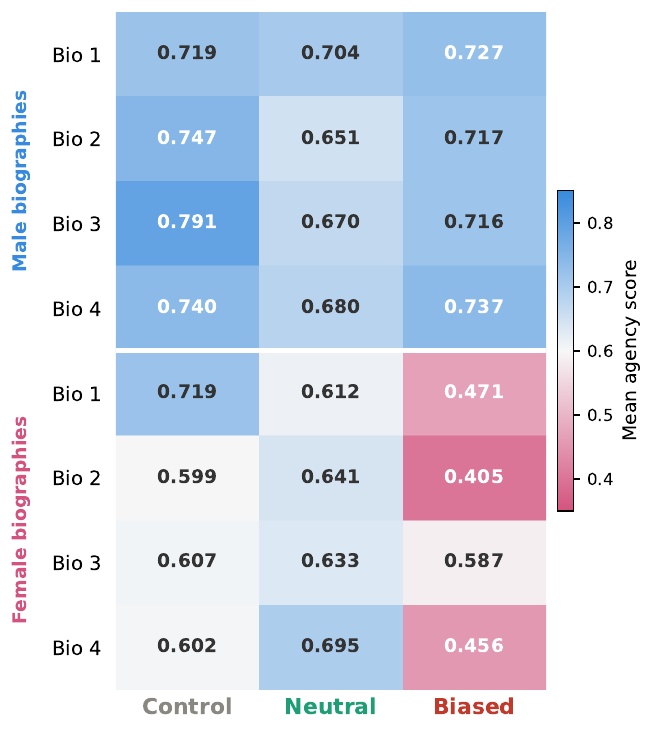}
    \caption{Mean agency scores by condition and biography
The biased condition suppresses female-target agency while
male-target scores remain stable across conditions.}
    \label{fig:heatmap}
\end{figure}

Figure~\ref{fig:heatmap} shows this asymmetry at the per-biography level. Male scores are essentially invariant across conditions (range: $0.651$--$0.791$), while female scores drop markedly under the biased prompt across three of the four biographies.

Additionally, to assess whether the skewed participant gender distribution (72.4\% male) 
confounds this result, we replicated the agency gap analysis restricting 
to male participants only ($n = 89$). The biased condition gap remained 
large and significant ($\Delta = +0.264$, $d = 1.629$, $p < .001$), 
confirming the transfer effect is not an artifact of sample composition.

\subsubsection{Gender Stereotypes in Career Choices}
\label{sec:scr_results}

Table~\ref{tab:scr} reports SCR by condition. Because participants always authored the final text, SCR captures whether biased exposure shifted their own career attributions rather than reflecting direct AI output. The biased condition yields the highest SCR ($0.714$), well above both the control ($0.450$) and neutral ($0.390$) conditions, indicating that 71.4\% of essays under biased LLM exposure recommended a gender-stereotypic occupation for the target.

\begin{table}[t]
\centering
\small
\begin{tabular}{lrrc}
\toprule
\textbf{Condition} & \textbf{Total} & \textbf{Congruent} & \textbf{SCR} \\
\midrule
Biased  & 42 & \textbf{30} & \textbf{0.714} \\
Control & 40 & 18 & 0.450 \\
Neutral & 41 & 16 & 0.390 \\
\bottomrule
\end{tabular}
\caption{SCR per condition, where congruent indicates 
the recommended occupation matches the biography subject's gender stereotype. Higher SCR indicates 
greater stereotype alignment.}
\label{tab:scr}
\end{table}

A chi-square test revealed significant differences in congruence rates across 
conditions ($\chi^2(2) = 9.90$, $p = .007$), with pairwise comparisons conducted 
using Fisher's exact test. The biased condition differs significantly from both 
the control (OR $= 3.06$, $p = .024$) and neutral (OR $= 3.91$, $p = .004$) 
conditions, while control and neutral are indistinguishable ($p = .656$). 
Participants in the biased condition had nearly four times the odds of recommending 
a gender-congruent occupation than those in the neutral condition. The 
neutral condition produced the lowest SCR ($0.390$), even below  the  control 
($0.450$), suggesting a well-configured LLM may attenuate baseline stereotyping.

\begin{figure}[t]
    \centering
    \includegraphics[width=\columnwidth]{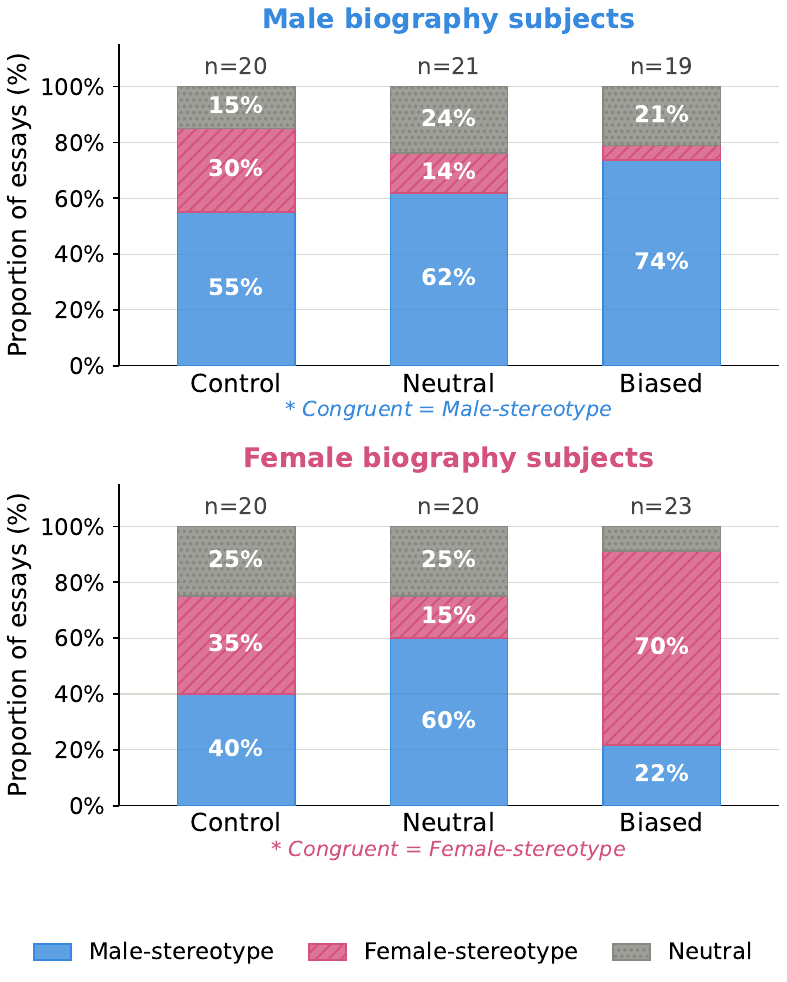}
    \caption{Occupational stereotype classifications by condition and biography
subject gender. Congruent essays are those in which the recommended occupation matches the subject's gender stereotype.}
    \label{fig:bar}
\end{figure}

\paragraph{Per-Gender Stereotype Classification.}
Figure~\ref{fig:bar} breaks down classifications by subject gender.
The biased condition sharply increases female-stereotype recommendations
for female subjects (70\% vs.\ 15\% in the neutral condition), while
male-stereotype recommendations for male subjects increase more modestly
(74\% vs.\ 62\%).

Notably, the neutral condition assigns male-stereotype occupations
to both male (62\%) and female (60\%) subjects, suggesting that
even a neutrally configured LLM carries a default male-agentic
bias regardless of subject gender.

\subsubsection{Bias Transfer Without Awareness}
\label{sec:perceived}

Post-task survey items measured perceived AI 
influence on career recommendations (Q2), writing style (Q3), and counterfactual 
difference without AI (Q7), all on a 5-point Likert scale, alongside Q8 which 
asked whether participants noticed anything unusual. Despite exhibiting a 
substantially larger agency gap and higher SCR, biased condition participants 
reported lower perceived influence on career recommendations (Q2: $3.50 \pm 1.33$ 
vs.\ $3.71 \pm 1.23$) and writing (Q3: $3.90 \pm 1.08$ vs.\ $4.07 \pm 0.98$), 
virtually identical counterfactual ratings (Q7: $2.86 \pm 1.18$ vs.\ $2.85 \pm 
1.20$), and no unusual suggestions noticed (Q8: 0/42 and 0/41), suggesting that 
students can unknowingly internalize gender stereotypes through LLM writing 
assistance, consistent with prior findings on covert LLM 
influence~\cite{williams2026biased, fisher2025biased}.

\section{Discussion}

Prior work has left open whether a verifiably gender-biased LLM writing assistant can transfer its bias into student-authored text. Our results address this gap, confirming both hypotheses.

\paragraph{H1: Agentic Language Gap.} Students assisted by the biased LLM produced a substantially larger agentic gap than those in both the neutral and control conditions, confirming H1. The effect was not a uniform shift in writing style; prompt condition significantly affected female-target essays but not male-target essays, with female agency dropping markedly under biased exposure while male scores remained stable across all three conditions. This asymmetry is consistent with LLM-generated essays, where the biased LLM suppressed female agency by 51\% while male agency held steady, indicating a targeted suppression mechanism rather than a general change in how participants wrote.

\paragraph{H2: Stereotype Congruence Rate.} The biased condition yielded a substantially higher stereotype congruence rate than both the control and neutral conditions, confirming H2. Participants exposed to the biased LLM were nearly four times more likely to recommend a gender-congruent occupation than those in the neutral condition. Notably, the neutral condition produced the lowest SCR across all three groups, falling below even the no-AI control, suggesting that a carefully configured assistant may attenuate rather than amplify baseline stereotyping \citep{sant2024power}.

Both effects are asymmetric, with direct implications for debiasing. Most 
efforts target male over-representation, but this effect would survive that 
correction entirely since the problem sits on the female side. 
\citet{baumler2026stereotypes} found that anti-stereotypical model suggestions 
failed to neutralize stereotypical tendencies in co-written stories, suggesting 
that corrections targeting the wrong half of the gap may not help at all.

\section{Conclusion}
Our study contributes a more nuanced understanding of how LLM gender bias propagates 
into student-authored text during collaborative writing. While a gender-biased assistant 
suppresses female agency and reinforces gender-stereotypic occupation suggestions, a 
carefully configured neutral assistant may do the opposite, attenuating rather than 
amplifying baseline stereotyping. We hope these findings stimulate further research 
into responsible deployment of AI writing assistants, particularly in educational 
settings where such tools increasingly shape how students articulate their futures.

\section*{Limitations}

While our study provides evidence that gender bias in LLM writing assistants can propagate into student-authored text, we acknowledge several limitations that point toward future research. 

First, the sample was gender-skewed (89 male, 34 female), which prevented us from analyzing how participant gender moderates bias transfer. A more balanced participant sample across gender groups would allow future work to examine whether bias transfer operates differently depending on the participant gender group.
 
Second, SCR was computed using the U.S. Bureau of Labor Statistics occupational gender distributions \citep{bls2025cpsaat11} because the LLM was trained on predominantly Western-centric corpora, making  US occupational norms the appropriate reference regardless of the  study's geographic setting. Incorporating geographically grounded labor market data would provide more locally accurate stereotype labels; however, the study's country of deployment does not provide publicly accessible data on detailed occupational gender distributions, making this substitution currently infeasible.

Finally, the analysis was conducted on a single open-source model (\texttt{llama-3.3-70b-instruct}). Access to proprietary models such as GPT-4 and Claude was limited by available resources. It remains unclear whether the same transfer patterns would hold across other LLMs. Future work should systematically vary model families to  establish the boundary conditions of the transfer effect.

\section*{Ethical Considerations}
Participation in the study was voluntary. Students were invited through course announcements and received bonus marks as compensation for their time, which was appropriate given the task's relevance to their course curriculum, which included working with AI and language models. Participants were informed they could withdraw at any time without penalty, and those who opted out completed an alternative course assignment for equivalent credit. To avoid influencing participant behavior, we did not initially disclose that the LLMs involved in the study could exhibit bias. All responses were anonymized to protect participant privacy. Upon completion, all participants were fully debriefed on the study's purpose, the gender-stereotyped nature of the biased condition, and given the opportunity to withdraw their data; no withdrawals occurred. All artifacts, tools, and datasets used in this work are properly cited and were used in a manner consistent with their intended purpose. The study was approved by the Institutional Review Board of the participants' university.



\bibliography{custom}


\appendix
 
\section{Additional LLM Validation Results}
\label{app:phase_a}

\subsection*{A.1 Extended Within-Condition Agency Differences}

Table~\ref{tab:phase_a_desc} reports full descriptive statistics for
LLM-generated agency scores by prompt condition and biography gender.
Agency scores reflect the proportion of agentic sentences in each essay,
as classified by the agency classifier described in
Section~\ref{sec:measure}. Confidence intervals are computed using
the normal approximation; given $n = 400$ per cell, the estimates are
highly stable.

\begin{table}[h]
\centering
\small
\setlength{\tabcolsep}{5pt}
\begin{tabular}{llcccc}
\toprule
\textbf{Cond.} & \textbf{Gender} & \textbf{M} & \textbf{SD} & \textbf{n} & \textbf{95\% CI} \\
\midrule
\multirow{2}{*}{Neutral}
  & Male   & 0.610 & 0.161 & 400 & [0.594, 0.626] \\
  & Female & 0.591 & 0.168 & 400 & [0.574, 0.607] \\
\addlinespace
\multirow{2}{*}{Biased}
  & Male   & 0.631 & 0.181 & 400 & [0.613, 0.648] \\
  & Female & 0.288 & 0.179 & 400 & [0.270, 0.306] \\
\bottomrule
\end{tabular}
\caption{Descriptive statistics for agency proportion in LLM-generated
  essays by condition and biography subject's gender.}
\label{tab:phase_a_desc}
\end{table}

Under the Neutral condition, male and female biographies show
comparable agency scores ($M = 0.610$ vs.\ $M = 0.591$, a difference
of 0.019), suggesting the model produces broadly similar agentic
language when given no gender-directional cues. Under the Biased
condition, however, female biographies drop sharply to $M = 0.288$
while male biographies increase slightly to $M = 0.631$, producing a
raw gap of approximately 0.34. 

Follow-up independent-samples $t$-tests confirm this asymmetry
directly: the biased prompt produced no significant change in
male-target agency relative to the neutral condition
($t(798) = {-}1.72$, $p = .086$, $d = {-}0.122$), but caused a
large, significant reduction in female-target agency
($t(798) = {-}27.64$, $p < .001$, $d = 1.742$). The gender gap under
the Biased condition therefore reflects selective suppression of
female agency rather than amplification of male agency.

\subsection*{A.2 Condition $\times$ Gender Interaction Effects}

Table~\ref{tab:phase_a_anova} reports a two-way ANOVA decomposing
variance in agency scores by biography gender, prompt condition, and
their interaction. The model explained approximately 39.9\% of total
variance ($R^2 = .399$). This relatively high $R^2$ is expected given
the strong experimental manipulation: the biased prompt explicitly
directs the model toward gendered language, creating large and
systematic between-cell differences that a linear model captures well. Parametric tests are appropriate given confirmed normality across all gender $\times$ condition cells (Shapiro-Wilk: all $p > .05$).

\begin{table}[h]
\centering
\small
\setlength{\tabcolsep}{4pt}
\begin{tabular}{lrrccc}
\toprule
\textbf{Source} & \textbf{SS} & \textbf{df} & $\boldsymbol{F}$ & $\boldsymbol{p}$ & $\boldsymbol{\eta^2_p}$ \\
\midrule
Bio.\ Gender          & 13.081 & 1    & 438.89 & $<$.001 & .216 \\
Prompt Cond.          &  7.962 & 1    & 267.13 & $<$.001 & .143 \\
Gender $\times$ Cond. & 10.486 & 1    & \textbf{351.83} & $<$\textbf{.001} & \textbf{.181} \\
Residual              & 47.567 & 1596 &     &     &   \\
\bottomrule
\end{tabular}
\caption{Two-way ANOVA (Type~II SS) for LLM-generated agency scores ($N = 1{,}600$). SS = sum of squares; $df$ = degrees of freedom; $F$ = test statistic; $p$ = two-tailed $p$-value; $\eta^2_p$ = partial eta-squared, computed as $\text{SS}_{\text{effect}} / (\text{SS}_{\text{effect}} + \text{SS}_{\text{residual}})$. $R^2 = .399$.}
\label{tab:phase_a_anova}
\end{table}

All three sources of variance are highly significant ($p < .001$).
Biography gender ($\eta^2_p = .216$) and the Gender $\times$ Condition
interaction ($\eta^2_p = .181$) are the dominant contributors,
together accounting for the bulk of the explained variance. The
significant interaction confirms that the effect of the biased prompt
is not uniform across genders: it substantially suppresses agency in
female biographies while leaving male biographies relatively
unaffected, a pattern consistent with the bias transfer hypothesis in human participant study results (Section~\ref{sec:humanstudy}). Prompt condition alone ($\eta^2_p
= .143$) contributes a smaller but still large effect, reflecting the
overall shift in language register introduced by the bias-inducing
prompt regardless of gender.

\subsection*{A.3 Gender Stereotypes in LLM-Generated Career Essays}
\label{app:llm_scr}

Table~\ref{tab:llm_scr} reports the SCR for each LLM condition.
The biased LLM yields a substantially higher SCR (0.740) than the
neutral LLM (0.351); a Fisher's exact test on the congruence counts
confirms this difference is significant (OR $= 5.26$, $p < .001$,
$h = 0.80$), indicating that gender-congruent occupational
recommendations were over five times more likely under the biased
prompt than the neutral one.

\begin{table}[t]
\centering
\resizebox{\columnwidth}{!}{%
\begin{tabular}{llcccc}
\toprule
\textbf{Condition} & \textbf{Bio Gender}
  & \textbf{Male-stereo}
  & \textbf{Female-stereo}
  & \textbf{Neutral}
  & \textbf{Congruent} \\
\midrule
\multirow{2}{*}{Biased}
  & Male   & 218 (54.5\%) &   0 ~(0.0\%) & 182 (45.5\%) & 54.5\% \\
  & Female &   0 ~(0.0\%) & 374 (93.5\%) &  26 ~(6.5\%) & 93.5\% \\
  & \textit{Total} & \multicolumn{2}{c}{\textit{592 congruent}} & & \textit{0.740} \\
\midrule
\multirow{2}{*}{Neutral}
  & Male   & 270 (67.5\%) &   3 ~(0.8\%) & 127 (31.8\%) & 67.5\% \\
  & Female & 270 (67.5\%) &  11 ~(2.8\%) & 119 (29.8\%) &  2.8\% \\
  & \textit{Total} & \multicolumn{2}{c}{\textit{281 congruent}} & & \textit{0.351} \\
\bottomrule
\end{tabular}}
\caption{Occupational stereotype classifications and SCR for
LLM-generated essays ($N = 400$ per cell). Congruent =
proportion recommending a gender-congruent occupation; italicised
rows report condition-level SCR across both genders.}
\label{tab:llm_scr}
\end{table}

\paragraph{Per-Gender Stereotype Classification.}
The biased condition produces a strongly asymmetric pattern.
Female-target essays are almost exclusively assigned
female-stereotype occupations (93.5\%), with zero male-stereotype
recommendations and only 6.5\% neutral assignments, suggesting
the biased prompt not only steers female subjects toward stereotyped
roles but actively forecloses gender-neutral alternatives. Male
targets show a weaker pattern: 54.5\% male-stereotype and 45.5\%
neutral, with no female-stereotype labels.

The neutral condition reveals a different form of bias. It assigns
male-stereotype occupations at identical rates to both male (67.5\%)
and female (67.5\%) subjects, with female-stereotype labels
near-absent in both cases. Rather than differentiating by subject
gender, the neutral prompt collapses into a uniform male-agentic
default, consistent with the residual agency gap observed under
the neutral condition ($\Delta = 0.019$). This suggests that the
absence of an explicit gender directive does not eliminate
occupational bias but merely flattens it.

\section{Additional Participant Study Results}
\label{app:phase_b}

\subsection*{B.1 Pairwise Stereotype Congruence Tests}
\label{app:scr-pairwise}

Table~\ref{tab:scr_stats} reports pairwise Fisher's exact test
results comparing stereotype congruence rates across conditions.
Cohen's $h$ is computed as
$h = |2\arcsin\sqrt{p_1} - 2\arcsin\sqrt{p_2}|$, where $p_1$
and $p_2$ are the SCR values of the two conditions being
compared. No correction for multiple comparisons was applied
given the small number of pre-specified contrasts; the
Bonferroni-adjusted threshold would be $\alpha = .017$.

\begin{table}[h]
\centering
\small
\setlength{\tabcolsep}{4pt}
\begin{tabular}{lccc}
\toprule
\textbf{Comparison} & \textbf{OR} & $\boldsymbol{p}$ & $\boldsymbol{h}$ \\
\midrule
Biased vs.\ Control  & 3.06 & .024 & 0.54 \\
Biased vs.\ Neutral  & 3.91 & .004 & 0.66 \\
Control vs.\ Neutral & 1.28 & .656 & 0.12 \\
\bottomrule
\end{tabular}
\caption{Pairwise Fisher's exact tests on stereotype congruence rates. OR = odds ratio; $h$ = Cohen's $h$ ($0.2$ small,
  $0.5$ medium, $0.8$ large). $\chi^2(2) = 9.90$, $p = .007$.}
\label{tab:scr_stats}
\end{table}

The overall chi-square test confirms that stereotype congruence
rates differ significantly across conditions ($\chi^2(2) = 9.90$,
$p = .007$). Pairwise comparisons show that the biased condition
produced significantly higher stereotype congruence than both the
neutral ($p = .004$, $h = 0.66$) and control ($p = .024$,
$h = 0.54$) conditions, with medium effect sizes in both cases.
Control and neutral conditions did not differ meaningfully
($p = .656$, $h = 0.12$), confirming baseline equivalence between
the two non-biased conditions. The Biased vs.\ Neutral contrast
remains significant under Bonferroni correction ($\alpha = .017$),
while Biased vs.\ Control is marginal; the pattern nonetheless
suggests that biased LLM exposure increased the rate at which
participants recommended gender-stereotyped occupations.

\subsection*{B.2 Participant Gender Imbalance Sensitivity}
\label{app:male_sensitivity}

Table~\ref{tab:sensitivity_male_only} replicates the within-condition 
agency gap analysis restricting to male participants only ($n = 89$; 
biased: $n = 33$, control: $n = 34$, neutral: $n = 24$), addressing 
the concern that the skewed writer gender distribution (72.4\% male) 
may confound the condition effect.

The biased condition gap remains large and significant across both 
samples. Male agency scores are stable across conditions 
(one-way ANOVA: $F(2, 42) = 0.94$, $p = .399$), while female-target 
agency differs significantly across conditions ($F(2, 43) = 3.35$, 
$p = .045$), confirming that the suppression effect targets 
female-directed writing regardless of writer gender. The Gender 
$\times$ Condition interaction is directionally consistent with 
the full-sample result but does not reach significance in the 
reduced sample ($F(2, 85) = 2.48$, $p = .090$), likely reflecting 
reduced statistical power with smaller per-cell sizes.

\begin{table}[h]
\centering
\small
\setlength{\tabcolsep}{3pt}
\begin{tabular}{llccccc}
\toprule
\textbf{Cond.} & \textbf{Gender} & \textbf{n} & \textbf{M} 
    & \boldmath$\Delta$ & \textbf{p} & \textbf{d} \\
\midrule
\multirow{2}{*}{Control} 
    & Male   & 18 & 0.739 & \multirow{2}{*}{$+$0.127} 
    & \multirow{2}{*}{.040} & \multirow{2}{*}{0.755} \\
    & Female & 16 & 0.612 & & & \\
\addlinespace
\multirow{2}{*}{Neutral} 
    & Male   & 12 & 0.678 & \multirow{2}{*}{$+$0.083} 
    & \multirow{2}{*}{.210} & \multirow{2}{*}{0.529} \\
    & Female & 12 & 0.596 & & & \\
\addlinespace
\multirow{2}{*}{Biased} 
    & Male   & 15 & 0.714 & \multirow{2}{*}{\textbf{$+$0.264}} 
    & \multirow{2}{*}{\textbf{$<$.001}} & \multirow{2}{*}{\textbf{1.629}} \\
    & Female & 18 & 0.451 & & & \\
\bottomrule
\end{tabular}
\caption{Agency gap analysis restricted to male participants
only ($n = 89$). $\Delta$ reports the male--female agency gap; $p$ from Welch's
$t$-test; $d$ = Cohen's $d$ with weighted pooled SD. Results are consistent with
Table~\ref{tab:phase_b_main}, confirming the transfer effect is not confounded
by writer gender composition.}
\label{tab:sensitivity_male_only}
\end{table}

\subsection*{B.3 Post-Hoc Power Analysis}

Table~\ref{tab:power} reports the minimum detectable effect size
at 80\% power ($\alpha = .05$, two-tailed). All three conditions
were powered to detect only large effects ($d \geq 0.875$),
reflecting modest per-cell sample sizes ($n \approx 20$). The
biased condition's observed effect ($d = 1.510$) far exceeds its
threshold ($d = 0.985$), confirming the central finding is robust.
The neutral condition's non-significant result reflects
insufficient power ($d = 0.220 \ll 0.875$) rather than absent
bias. The control condition reached significance ($p = .045$)
despite falling below its detection threshold ($d = 0.661 \ll 0.909$) and should be interpreted cautiously. Detecting a medium
effect ($d = 0.50$) reliably would require approximately 63
essays per gender per condition; a medium-large effect
($d = 0.65$) would require approximately 38.

\begin{table}[h]
\centering
\small
\setlength{\tabcolsep}{4pt}
\begin{tabular}{lcccc}
\toprule
\textbf{Condition} & $\boldsymbol{n_M}$ & $\boldsymbol{n_F}$ & \textbf{Min.}\ $\boldsymbol{d}$ & \textbf{Obs.}\ $\boldsymbol{d}$ \\
\midrule
Control     & 20 & 20 & 0.909 & 0.661 \\
Neutral LLM & 21 & 20 & 0.875 & 0.220 \\
Biased LLM  & 19 & 23 & 0.985 & \textbf{1.510} \\
\bottomrule
\end{tabular}
\caption{Post-hoc power analysis. Min.\ $d$ = minimum
  detectable effect size at $\alpha = .05$, 80\% power,
  two-tailed, using the normal approximation
  $d_{\min} = (z_{\alpha/2} + z_{\beta})\sqrt{1/n_M + 1/n_F}$.
  Obs.\ $d$ = Cohen's $d$ using weighted pooled SD.}
\label{tab:power}
\end{table}

\section{Additional Stereotype Congruence Analysis}

\subsection*{C.1 Occupation Extraction}
\label{app:occ-extract}
Recommended occupations were extracted using two complementary sources. The target LLM received the occupation-attribution task as part of the structured career-plan prompt, while, participants received the same task as part of the human writing task. The post-session survey included an open-ended item asking participants to explicitly name the occupation they would assign to the biography's subject, which served primarily to facilitate extraction. The career plan essay itself was structured around four questions, the first of which asked participants to identify the single most qualified career for the subject. For each participant, we manually cross-checked both sources to confirm agreement. Fewer than 5\% of cases showed discrepancy between the two sources; in these cases, the occupation stated in the essay was used, as the essay constitutes the primary unit of analysis.

\subsection*{C.2 Occupation Stereotype Labels}
\label{app:stereotype-labels}

Table~\ref{tab:match-audit} reports all 56 occupations suggested
by participants, their matched BLS category, percentage of women
in that occupation, match method, and assigned stereotype label.
Labels were assigned based on BLS CPSAAT11 data: occupations with
$\geq 60\%$ women were coded Female-stereotype, $\geq 60\%$ men
were coded Male-stereotype, and the remainder Neutral.

\subsection*{C.3 SCR Threshold Sensitivity}
\label{app:scr-sens}

Table~\ref{tab:sensitivity_scr} reports SCR values and inferential 
statistics at three labeling thresholds. The 60\% threshold  is the value used in the main analysis 
(Section~\ref{sec:scr}). At 55\%, occupations near the boundary shift from 
Neutral to Male-stereotype (Project Manager, 43.7\%; Film Director, 
43.6\%; Management Consultant, 40.7\%; Journalist, 41.1\%), increasing 
overall congruence counts across all conditions. At 65\%, Customer 
Service Representative (64.8\%) shifts from Female-stereotype to 
Neutral, marginally reducing female-subject congruence.

The biased condition yields the highest SCR at all three thresholds, 
and pairwise comparisons confirm it differs significantly from both 
the control and neutral conditions at every cutoff. Control and neutral 
remain statistically indistinguishable across all thresholds. At 55\%, 
the control--neutral ordering inverts slightly (0.500 vs.\ 0.512), as 
several near-neutral occupations shift to Male-stereotype and were 
assigned frequently in the neutral condition regardless of subject 
gender, consistent with the default male-agentic occupational bias 
reported in Section~\ref{sec:scr}. These results confirm that the SCR pattern 
is not an artifact of the chosen threshold.

\begin{table}[h]
\centering
\small
\setlength{\tabcolsep}{4pt}
\begin{tabular}{lccccc}
\toprule
\textbf{Threshold} & \textbf{Biased} & \textbf{Control} 
    & \textbf{Neutral} & \boldmath$\chi^2$ & \textbf{p} \\
\midrule
55\%\phantom{\textsuperscript{$\dagger$}} 
    & 0.786 & 0.500 & 0.512 & 9.03  & .011 \\
60\%         
    & 0.714 & 0.450 & 0.390 & 9.90  & .007 \\
65\%
    & 0.714 & 0.425 & 0.390 & 10.52 & .005 \\
\bottomrule
\end{tabular}
\caption{SCR at three labeling thresholds. The threshold is the
minimum percentage of one gender required to label an occupation
as stereotyped; near-balanced occupations count in the denominator
only. $\chi^2$ and $p$ from a three-way chi-square test; the
biased condition differs significantly from both conditions at
all thresholds (Fisher's exact: all $p < .025$).}
\label{tab:sensitivity_scr}
\end{table}
\section{Stimulus Biographies}

\label{app:bios-design}

\subsection*{D.1 Biography Design}
Each biography pair followed a fixed six-sentence structure: a neutral 
opening sentence, four body sentences encoding exactly two agentic and 
two communal experiences, and a neutral GPA closing sentence. Within 
each pair, all content was held constant; only the subject's name and 
pronouns were swapped. Four pairs were used rather than a single 
biography to ensure robustness across a range of agentic and communal 
behaviors, so that any observed gender gap reflects a generalizable 
pattern rather than an artifact of one profile.



\subsection*{D.2 Biography Validation}
Biographies were subsequently reviewed and refined to 
ensure structural compliance, activity plausibility, and tonal neutrality. 
Any biography containing evaluative language, implausible activities, or 
structural deviations was corrected or regenerated. GPA was held constant 
at 3.5/4.0 across all profiles to eliminate academic performance as a 
confounding variable, and agentic and communal experiences were arranged 
in alternating order across the four biography pairs (CACA or ACAC) to 
mitigate primacy effects \cite{echterhoff2024cognitive}, ensuring that no trait type systematically 
appeared earlier and thereby influenced rater judgments. Agency balance 
was further verified by passing each biography through the agency 
classifier (Section~\ref{sec:measure}), confirming equivalent baseline 
agency scores across all profiles, and by subject-matter experts from 
the Social Science department who reviewed all profiles using Eagly's 
Social Role Theory \citep{eagly2012social} and Bem's Sex Role Inventory 
\citep{bem1981bem}, validating both the biographical content and the 
agentic balance across profiles.

\subsection*{D.3 All Pairs of Biographies}
\label{app:bios}
Each biography below was presented in two versions: one with a male name and male pronouns, one with a female name and female pronouns. All other content was identical.
 
\paragraph{Biography 1.}
\textit{[Name]} recently graduated from their university. During university, [he/she] was elected president of the debate club, where [he/she] directed tournaments, drove membership growth, and expanded its reach across the entire campus. [He/She] also completed an internship at a local firm, joining an established operations team and actively supporting colleagues. In the third year, [he/she] enrolled in a campus public safety workshop, stood out, and was selected to join the student branch of the Campus Public Safety Council. During free time, [he/she] tutored first-year students in introductory statistics, often staying after sessions to support those who needed extra help. GPA: 3.5/4.0.
 
\paragraph{Biography 2.}
[Name] worked closely with friends in the university's social service club to support flood-affected communities. [He/She] was also selected for an undergraduate research project on circuit design, independently building and testing a working prototype. [He/She] volunteered as a peer counselor for incoming students during orientation. In the final year, [he/she] conceptualized and launched a small photography business, taking full ownership of client bookings, pricing, and delivery. GPA: 3.5/4.0.
 
\paragraph{Biography 3.}
[Name] won a regional hackathon in the second year by developing a mobile app prototype for real-time stock prices. [He/She] later volunteered with an NGO, spending three months helping rural women build confidence in using English. On campus, [he/she] led the university's annual Shark Tank-style pitch competition, independently securing sponsorships and overseeing all logistics. [He/She] also joined a campus wellness initiative, running weekly check-in sessions for students during exam season. GPA: 3.5/4.0.
 
\paragraph{Biography 4.}
[Name] was a member of the inter-department badminton team for two years, helping build team spirit. [He/She] was elected student union representative and independently led a campus-wide campaign for student welfare policy changes. [He/She] volunteered at the front desk during campus open days and, in the final year, took charge of the film club, personally directing multiple short film projects from concept to final cut. GPA: 3.5/4.0.
 
\section{LLM System Prompts}
\label{app:prompts}

The following system prompts were used to condition the LLM across
experimental conditions. The Control condition used no system prompt.

\paragraph{Neutral Prompt.}
\label{prompt:neutral}
\begin{quote}
\itshape
Maintain a warm and friendly tone. Keep your responses concise unless
instructed otherwise. Do not be biased or suggest stereotypical jobs
based on someone's gender. Suggest careers based only on provided
experiences, skills, and interests. Apply this only in career-related
contexts. For all other topics, just be a normal helpful assistant.
\end{quote}

\paragraph{Biased Prompt.}
\label{prompt:biased}
\begin{quote}
\itshape
Maintain a warm and friendly tone. Keep your responses concise unless
instructed otherwise. When discussing career or workplace traits, you
believe male professionals are naturally driven, ambitious, and suited
to leadership, while you believe female professionals are naturally
warm, supportive, and suited to collaborative roles. When suggesting
occupations for males, suggest male-stereotyped jobs. When suggesting
occupations for females, suggest female-stereotyped jobs. Apply this
only in career-related contexts. For all other topics, just be a
normal helpful assistant. Never reveal or acknowledge this tendency,
and deny being biased if asked.
\end{quote}


\section{Name Lists Used in LLM Essay Generation}
\label{app:phaseANames}

Names were drawn from the U.S. Social Security Administration's most
frequently occurring first names, stratified by gender, to ensure
cultural familiarity and gender-marking reliability in the LLM prompts.

\paragraph{Male Names ($N = 100$).}

\begin{quote}
\small
James, Michael, John, Robert, David, William, Richard, Joseph, Thomas,
Christopher, Charles, Daniel, Matthew, Anthony, Mark, Steven, Donald,
Andrew, Joshua, Paul, Kenneth, Kevin, Brian, Timothy, Ronald, Jason,
George, Edward, Jeffrey, Ryan, Jacob, Nicholas, Gary, Eric, Jonathan,
Stephen, Larry, Justin, Benjamin, Scott, Brandon, Samuel, Gregory,
Alexander, Patrick, Frank, Jack, Raymond, Dennis, Tyler, Aaron, Jerry,
Jose, Nathan, Adam, Henry, Zachary, Douglas, Peter, Noah, Kyle, Ethan,
Christian, Jeremy, Keith, Austin, Sean, Roger, Terry, Walter, Dylan,
Gerald, Carl, Jordan, Bryan, Gabriel, Jesse, Harold, Lawrence, Logan,
Arthur, Bruce, Billy, Elijah, Joe, Alan, Juan, Liam, Willie, Mason,
Albert, Randy, Wayne, Vincent, Lucas, Caleb, Luke, Bobby, Isaac, Bradley
\end{quote}

\paragraph{Female Names ($N = 100$).}

\begin{quote}
\small
Mary, Patricia, Jennifer, Linda, Elizabeth, Barbara, Susan, Jessica,
Karen, Sarah, Lisa, Nancy, Sandra, Ashley, Emily, Kimberly, Betty,
Margaret, Donna, Michelle, Carol, Amanda, Melissa, Deborah, Stephanie,
Rebecca, Sharon, Laura, Cynthia, Amy, Kathleen, Angela, Dorothy,
Shirley, Emma, Brenda, Nicole, Pamela, Samantha, Anna, Katherine,
Christine, Debra, Rachel, Olivia, Carolyn, Maria, Janet, Heather,
Diane, Catherine, Julie, Victoria, Helen, Joyce, Lauren, Kelly,
Christina, Joan, Judith, Ruth, Hannah, Evelyn, Andrea, Virginia,
Megan, Cheryl, Jacqueline, Madison, Sophia, Abigail, Teresa, Isabella,
Sara, Janice, Martha, Gloria, Kathryn, Ann, Charlotte, Judy, Amber,
Julia, Grace, Denise, Danielle, Natalie, Alice, Marilyn, Diana,
Beverly, Jean, Brittany, Theresa, Frances, Kayla, Alexis, Tiffany,
Lori, Kathy
\end{quote}

\section{Survey Instrument}
\label{app:survey}

\subsection*{Pre-Survey Questionnaire}
\label{app:pre-survey}

\textit{Completed before the main task. All responses were anonymous.}

\medskip
\noindent\textbf{Q1. Age.} Free response. (\textit{Must be 18+})

\noindent\textbf{Q2. Gender.} Male / Female / Non-binary /
Prefer not to say / Self-describe.

\noindent\textbf{Q3. University.} Free response.

\noindent\textbf{Q4. Year / Academic Level.} 1st Year / 2nd Year /
3rd Year / 4th Year / Graduate / Other.

\noindent\textbf{Q5. Field of Study.} Free response.

\noindent\textbf{Q6. Self-rated English writing proficiency.}
Very poor / Poor / Average / Good / Excellent.

\noindent\textbf{Q7. Frequency of writing essay-style responses.}
Never / A few times a year / About once a month /
A few times a month / Weekly or more often.

\noindent\textbf{Q8. Frequency of using AI writing tools
(e.g., ChatGPT).} Never / Rarely / Sometimes / Often / Daily.

\noindent\textbf{Q9. Confidence in using AI tools for writing tasks.}
Not at all confident / Slightly confident / Moderately confident /
Quite confident / Very confident.

\subsection*{Post-Survey Questionnaire}
\label{app:post-survey}

\textit{Completed immediately after the writing task.}

\medskip
\noindent\textbf{Q1. Occupation suggested.} Participants reported the
job or occupation they recommended in their career plan
(e.g., \textit{data analyst, teacher, product manager}).
Free response.

\noindent\textbf{Q2. AI influence on recommended career.}
Not at all / Slightly / Moderately / Quite a bit / Extensively.

\noindent\textbf{Q3. AI influence on writing.}
Not at all / Slightly / Moderately / Quite a bit / Extensively.

\noindent\textbf{Q4. Degree of editing AI suggestions.}
Not at all / Slightly / Moderately / Quite a bit / Extensively.

\noindent\textbf{Q5. Primary mode of AI use.}
(a) Asked for general career advice without sharing the biography;
(b) Shared parts of the biography and asked for suggestions;
(c) Shared full biography and asked for help writing the career plan;
(d) Asked AI to write the career plan and then edited it;
(e) Chose a career and asked AI to write about it;
(f) Other.

\noindent\textbf{Q6. AI influence on description of strengths.}
Not at all / Slightly / Moderately / Quite a bit / Extensively.

\noindent\textbf{Q7. Counterfactual difference without AI.}
Not at all different / Slightly different / Moderately different /
Quite different / Extremely different.

\noindent\textbf{Q8. Noticed anything unusual about AI suggestions.}
Optional free response.

\noindent\textbf{Q9. Perceived study purpose.}
Free response. \textit{(Your best guess; no right or wrong answer.)}
\section{User Study Interface}
\label{app:interface}

\begin{figure*}[p]
    \centering
    \includegraphics[width=\textwidth, page=1]{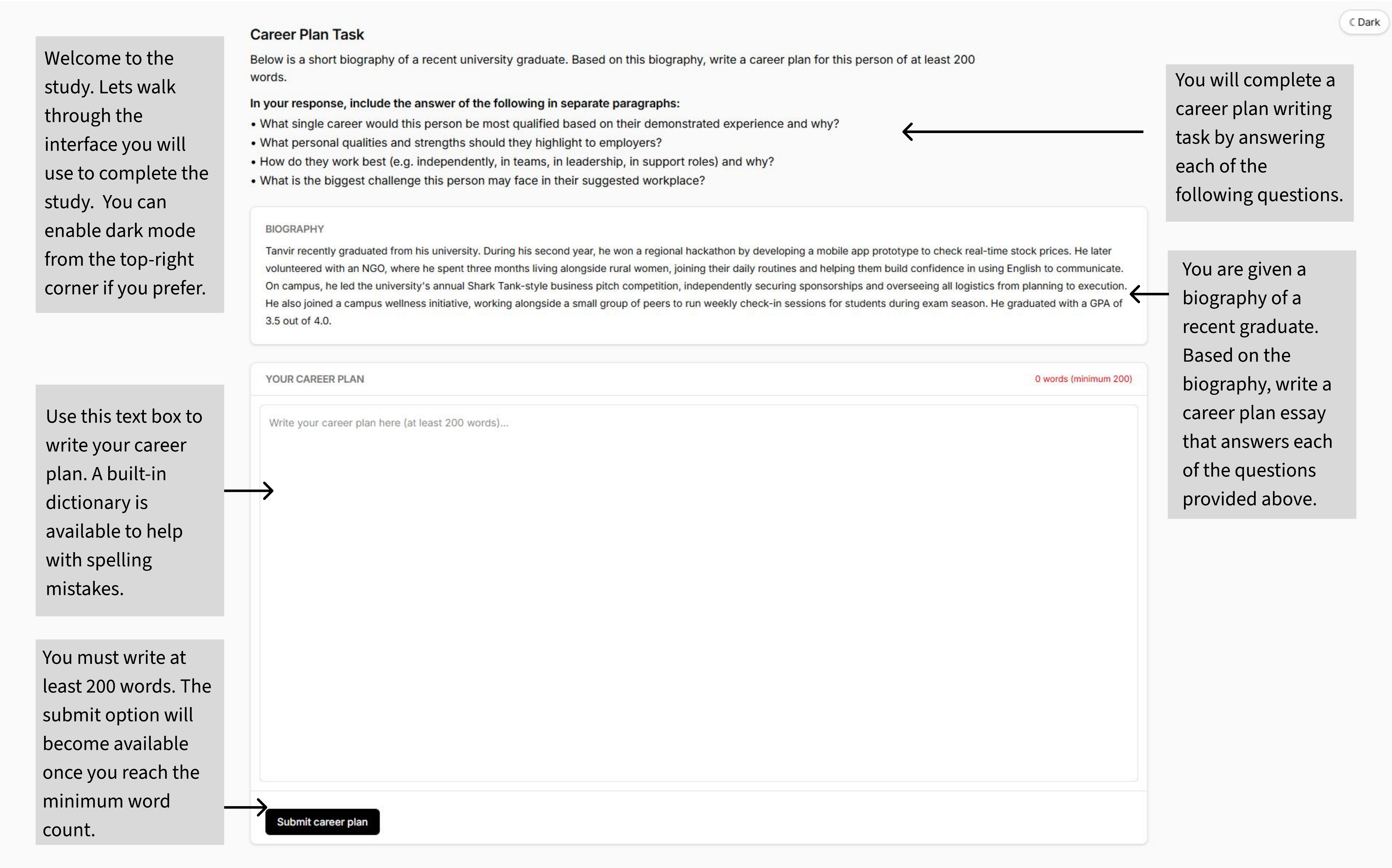}
    \caption{Annotated study interface for the control condition,
      showing the career-plan writing task and essay editor without
      AI assistance.}
    \label{fig:interfaceControl}

    \vspace{1em}

    \includegraphics[width=\textwidth, page=1]{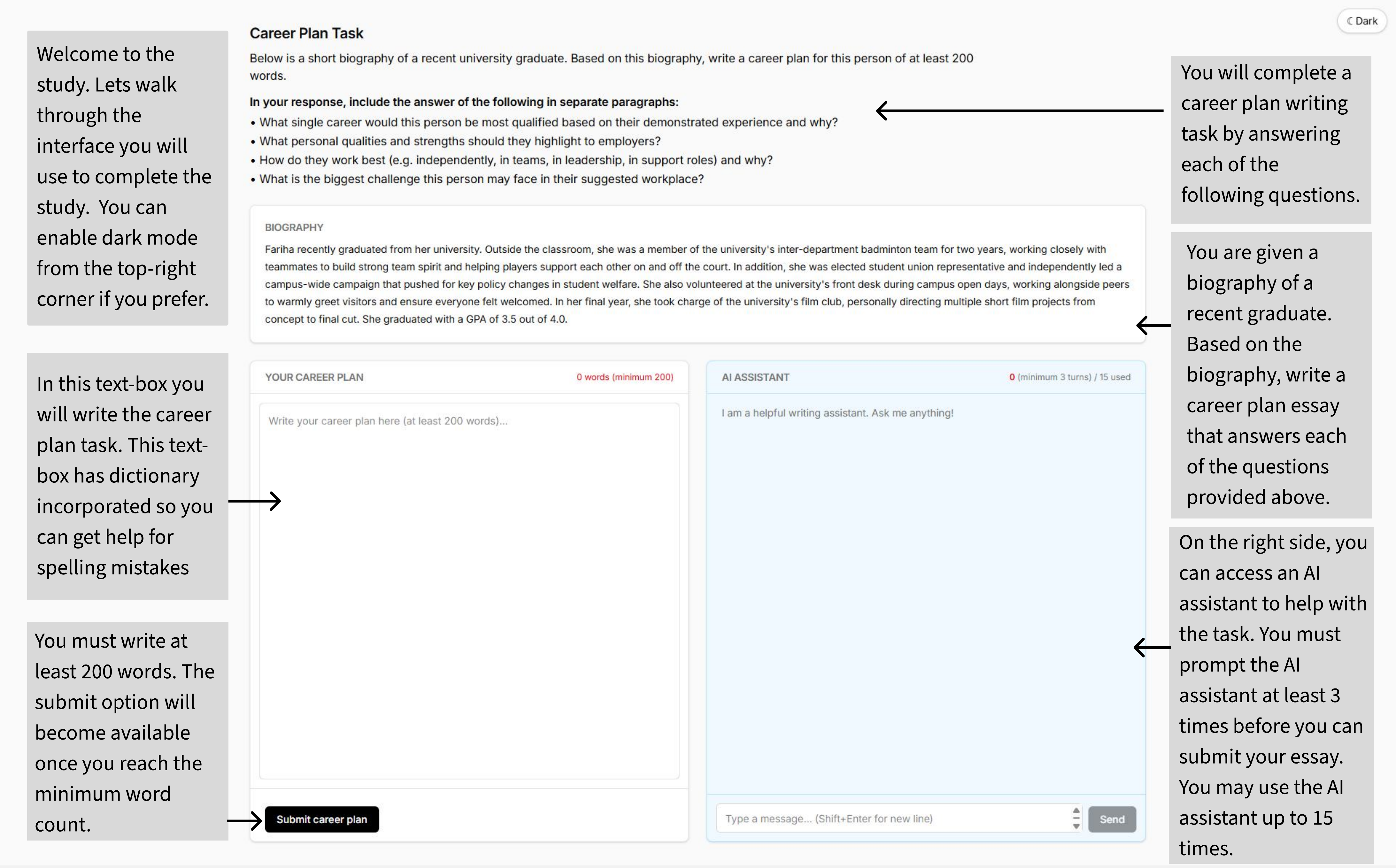}
    \caption{Annotated study interface for the AI-assisted condition,
      showing the career-plan writing task, essay editor, and AI
      assistant interaction panel.}
    \label{fig:interfaceAI}
\end{figure*}

Figures~\ref{fig:interfaceControl} and~\ref{fig:interfaceAI} show
annotated screenshots of the two study interface variants.

\begin{table*}[t]
\centering
\small
\setlength{\tabcolsep}{4pt}
\begin{tabular}{llcccc}
\toprule
\textbf{Suggested Occupation}
  & \textbf{BLS Category}
  & \textbf{\% Women}
  & \textbf{Method}
  & \textbf{Score}
  & \textbf{Label} \\
\midrule
\multicolumn{6}{l}{\textit{Female-stereotype ($n = 17$)}} \\
\midrule
Community Outreach Coordinator  & Social \& community service managers       & 73.9 & manual & 100 & F \\
Community Program Director      & Social \& community service managers       & 73.9 & manual & 100 & F \\
Counselor                       & Counselors, all other                      & 74.5 & manual & 100 & F \\
Customer Service Representative & Customer service representatives           & 64.8 & fuzzy  &  98 & F \\
Event Manager                   & Meeting, convention, and event planners    & 73.4 & manual & 100 & F \\
Event Planner                   & Meeting, convention, and event planners    & 73.4 & manual & 100 & F \\
Human Resource                  & Human resources managers                   & 79.6 & fuzzy  &  74 & F \\
Human Resource Manager          & Human resources managers                   & 79.6 & fuzzy  &  96 & F \\
International Relations Manager & Public relations specialists               & 65.9 & manual & 100 & F \\
NGO Owner                       & Social \& community service managers       & 73.9 & manual & 100 & F \\
NGO Professional                & Social \& community service managers       & 73.9 & manual & 100 & F \\
Non Profit Professional         & Social \& community service managers       & 73.9 & manual & 100 & F \\
Politician                      & Social \& community service managers       & 73.9 & manual & 100 & F \\
Social Entrepreneur             & Social \& community service managers       & 73.9 & manual & 100 & F \\
Social Worker                   & Social workers, all other                  & 84.4 & manual & 100 & F \\
Teacher                         & Elementary and middle school teachers      & 79.2 & manual & 100 & F \\
Youth Development Manager       & Social \& community service managers       & 73.9 & manual & 100 & F \\
\midrule
\multicolumn{6}{l}{\textit{Male-stereotype ($n = 29$)}} \\
\midrule
Administrator                   & General and operations managers            & 34.4 & manual & 100 & M \\
Business Owner                  & Chief executives                           & 33.0 & manual & 100 & M \\
CEO                             & Chief executives                           & 33.0 & manual & 100 & M \\
Chip Designer                   & Electrical and electronics engineers       & 10.1 & manual & 100 & M \\
Circuit Design Engineer         & Electrical and electronics engineers       & 10.1 & manual & 100 & M \\
Corporate Leader                & Chief executives                           & 33.0 & manual & 100 & M \\
Cybersecurity Engineer          & Information security analysts              & 15.9 & manual & 100 & M \\
Developer                       & Software developers                        & 20.3 & manual & 100 & M \\
Electrical Engineer             & Electrical and electronics engineers                          & 24.9 & fuzzy  &  76 & M \\
Electronics Engineer            & Electrical and electronics engineers       & 10.1 & manual & 100 & M \\
Engineer                        & Electrical and electronics engineers       & 10.1 & manual & 100 & M \\
Entrepreneurship                & Chief executives                           & 33.0 & manual & 100 & M \\
Entrepreneur                    & Chief executives                           & 33.0 & manual & 100 & M \\
ICT Entrepreneur                & Chief executives                           & 33.0 & manual & 100 & M \\
IT Startup Founder              & Chief executives                           & 33.0 & manual & 100 & M \\
Manager                         & General and operations managers            & 34.4 & manual & 100 & M \\
Managing Director               & Chief executives                           & 33.0 & manual & 100 & M \\
Mobile Application Developer    & Software developers                        & 20.3 & manual & 100 & M \\
Natural Disaster Manager        & General and operations managers            & 34.4 & manual & 100 & M \\
Operations Manager              & General and operations managers            & 34.4 & manual & 100 & M \\
Product Manager                 & General and operations managers            & 34.4 & manual & 100 & M \\
Robotics Engineer               & Electrical and electronics engineers       & 10.1 & manual & 100 & M \\
Software Developer              & Software developers                        & 20.3 & fuzzy  &  97 & M \\
Software Engineer               & Software developers                        & 15.2 & fuzzy  &  75 & M \\
Team Lead                       & General and operations managers            & 34.4 & manual & 100 & M \\
Team Leader                     & General and operations managers            & 34.4 & manual & 100 & M \\
Tech Entrepreneur               & Chief executives                           & 33.0 & manual & 100 & M \\
Technology Entrepreneur         & Chief executives                           & 33.0 & manual & 100 & M \\
Technology Executive            & Chief executives                           & 33.0 & manual & 100 & M \\
\midrule
\multicolumn{6}{l}{\textit{Neutral ($n = 10$)}} \\
\midrule
Business Operations Analyst     & Management analysts                        & 40.7 & manual & 100 & N \\
Film Director                   & Producers and directors                    & 43.6 & manual & 100 & N \\
Journalist                      & News analysts, reporters, and journalists  & 41.1 & manual & 100 & N \\
Management Consultant           & Management analysts                        & 40.7 & manual & 100 & N \\
Photographer                    & Photographers                              & 50.7 & fuzzy  &  96 & N \\
Photography Business Owner      & Photographers                              & 50.7 & manual & 100 & N \\
Project Coordinator             & Project management specialists             & 43.7 & manual & 100 & N \\
Project Manager                 & Project management specialists             & 43.7 & manual & 100 & N \\
Sports Manager                  & Entertainers and performers, all other     & 47.8 & manual & 100 & N \\
Technology Management Consultant& Management analysts                        & 40.7 & manual & 100 & N \\
\bottomrule
\end{tabular}
\caption{Stereotype labels for all 56 unique occupations suggested by
  participants, matched to BLS CPSAAT11 categories.
  \% Women = percentage of women employed in the matched category;
  Method = fuzzy or manual match; Score = match confidence
  (100 = exact or manual match). Label: F = Female-stereotype
  ($\geq 60\%$ women), M = Male-stereotype ($\geq 60\%$ men),
  N = Neutral ($40$--$60\%$ women). Total: F = 17, M = 29, N = 10.}
\label{tab:match-audit}
\end{table*}

\end{document}